\documentclass{article}

\usepackage{PRIMEarxiv}
\usepackage{float}
\usepackage[utf8]{inputenc} 
\usepackage[T1]{fontenc}    
\usepackage[colorlinks=true]{hyperref}
\usepackage{url}            
\usepackage{booktabs}       
\usepackage{amsfonts}       
\usepackage{nicefrac}       
\usepackage{microtype}      
\usepackage{lipsum}
\usepackage{fancyhdr}       
\usepackage{graphicx}       
\usepackage{xcolor}
\usepackage{graphicx}
\usepackage{amsmath}
\usepackage{amssymb}
\usepackage{booktabs}
\usepackage{multirow}
\usepackage{tabularray}
\usepackage{array}
\usepackage{xspace}

\makeatletter
\DeclareRobustCommand\onedot{\futurelet\@let@token\@onedot}
\def\@onedot{\ifx\@let@token.\else.\null\fi\xspace}

\makeatother
\usepackage{amsmath}
\usepackage{amssymb}
\usepackage{mathtools}
\usepackage{amsthm}
\DeclareMathOperator*{\argmax}{argmax}
\DeclareMathOperator*{\argmin}{argmin}

\title{Contour Completion using Deep Structural Priors}

\author{
    Ali Shiraee$^{\dagger}$\\
    Department of Computer Science \\
    Kharazmi University
   \And
    Morteza Rezanejad$^{\dagger}$\\
    Department of Psychology\\
    University of Toronto 
    \thanks{Dr. Morteza Rezanejad contributed to this article in his personal capacity as an adjunct researcher at the University of Toronto.
    \newline
 \indent ~$^{\dagger}$These authors made equal contributions to this article.\newline
\indent\indent Corresponding author: Morteza Rezanejad (\url{morteza.rezanejad@utoronto.ca}).
}
\\
    \And
    Mohammad Khodadad\\
    Faculty of Engineering\\
    McMaster University \\
   \And
    Dirk B. Walther\\
    Department of Psychology\\
    University of Toronto \\
   \And
    Hamidreza Mahyar\\
    Faculty of Engineering\\
    McMaster University
}
\begin{document}
\maketitle

\begin{abstract}
Humans can easily perceive illusory contours and complete missing forms in fragmented shapes. This work investigates whether such capability can arise in convolutional neural networks (CNNs) using deep structural priors computed directly from images. In this work, we present a framework that completes disconnected contours and connects fragmented lines and curves. In our framework, we propose a model that does not even need to know which regions of the contour are eliminated. We introduce an iterative process that completes an incomplete image and we propose novel measures that guide this to find regions it needs to complete. Our model trains on a single image and fills in the contours with no additional training data. Our work builds a robust framework to achieve contour completion using deep structural priors and extensively investigate how such a model could be implemented.
\end{abstract}


\section{Introduction}
\label{intro}

The human visual system is used to seeing incomplete outlines. Our brains can effortlessly group visual elements and fragmented contours that seem to be connected to each other. This power enables us to make shapes, organize disconnected visual features, and even properties of 3D surfaces when projected on 2D planes. 
\cite{rensink1998early} demonstrated how early vision may quickly complete partially-occluded objects using monocular signals. 
This capability of perceptual grouping has been studied in vision science for decades \cite{witkin1983role, lowe1985visual, sarkar1993perceptual, goodnow_childrens_1980}. Although there has been some work on perceptual grouping \cite{rezanejad2020medial,levinshtein2013multiscale, lee2015learning, rezanejad2019gestalt} in the past couple of years, it has been less studied in the past decade due to the enormous progress of deep neural networks and their success in dealing with the pixel-by-pixel inference of images.

Different types of lines and curves have been studied to maximize the connectivity of two broken ends in the planer contour completion problem \cite{rutkowski1979shape, ullman1976filling, brady1980shape}. Different types of lines and curves have been studied to maximize the connectivity of two broken ends in the planer contour completion problem. Geometry-based constraints can be utilized to address some challenges of contour completion problems, such as smoothness and curvature consistency \cite{horn1983curve, mumford1994elastica, sharon2000completion, mio2006contour}. However, such approaches only work for simple, smooth contours and usually fail in more complex settings.

On the other hand, we currently have deep models that could easily take an incomplete image and complete the missing regions using enough training data \cite{yu2018generative, elharrouss2020image, qin2021image}. The amazing capability of such models especially those that are trained on different modalities with millions or billions of training data \cite{brown2020language,ramesh2022hierarchical, saharia2022photorealistic}  raises the question of whether we need such a large amount of training to perceive all the visual cues that are present in an image, which underlies visual perception by humans. 
In human vision, Gestalt psychology \cite{koffka1922perception} suggests that our brain is designed to perceive structures and patterns that are grouped by some known rules. 
In this work, we show that some perceptual structures can also be learned from the image itself directly using architectures that enable such learning. Earlier work has shown that some forms of perceptual grouping can be achieved using computational models, such as stochastic completion fields \cite{williams1996local,williams1997local,DBLP:conf/bmvc/RezanejadGGMWGW21}. This type of learning resonates with some of the Gestalt perceptual grouping principles including ``proximity'', ``good continuation'' and ``similarity''. In scenarios where color and/or texture are present, the cue of ``similarity'' helps us group regions with consistent patterns \cite{camaro2020appearance}. When color and texture are present, they provide a collection of rich information for such cues. In the present article, we probe convolutional neural networks in a scenario where both are absent, and the neural network is dealing with just forms and shapes. 

Specifically, we explore whether the convolutional neural network architecture itself can give rise to some of these grouping cues when they are fed just contours and shapes alone. For years, neural networks have been treated as black boxes that can generalize images very well to multiple classes when there are enough training exemplars. One of the reasons that neural networks are trained on many exemplars is to avoid the problem of overfitting. On the other hand, we know that CNNs that generalize well to large classes of exemplars can easily overfit when those class labels are randomly permuted \cite{zhang2021understanding}. Inspired by this observation, \cite{ulyanov2018deep} suggest that image priors can be learned to a large extent through a generator network architecture \cite{goodfellow2020generative} that is solely trained on a single image. This encouraged us to take a deeper look at what structural information can be learned from a single-shape image and whether we can reconstruct some of those perceptual grouping capabilities using a generator network.

Inspired by \cite{zhang2021understanding,ulyanov2018deep}, in this work, we adopt a novel training regime to complete shapes and contours where we use a UNet architecture \cite{ronneberger2015u} with random initial weights and try to complete the contours within a single image without any training data. In our model, we imagine that the input image (i.e., the only image used to update the model's weights) is an image of fragmented contours. In this work, instead of training the model on multiple images fetched from a big image dataset, we imagine a random fixed tensor noise image as input to this model. At each iteration, the random noise tensor is inferred through our generative network and the network produces an outcome image. We introduce a novel loss function that enables this network to complete contours. This process repeats, and the weights of our network are updated gradually based on this loss function, which is an energy term defined based on the input image and the output of the network. 

\begin{figure}[!t]
	\centering
\begin{tabular}{cc}
    \multicolumn{2}{c}{\includegraphics[width = 0.4\textwidth]{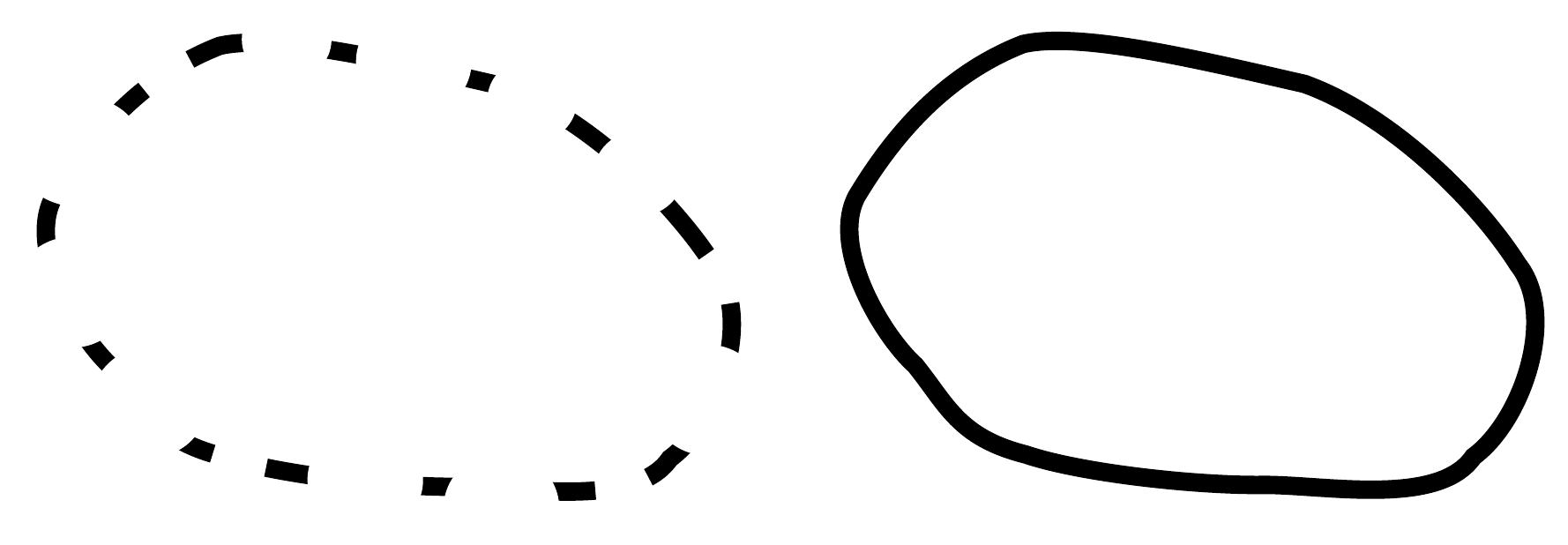}} \\
    \hspace{15mm} \textbf{(a)} & \hspace{15mm} \textbf{(b)}
\end{tabular}

	\label{figure_kaniza}
	\caption{Just by looking at subfigure \textbf{(a)}, we, as humans, can easily perceive a shape like the one in subfigure \textbf{(b)}. This is an extraordinary capability of our human brain and in this paper, we tried to see whether convolutional neural networks can show such capabilities. }
\end{figure}

The model will reconstruct the missing structures i.e., group fragmented contours that perceptually seem to be connected, before it fully overfits to the incomplete input image. 
Contributions of our work are summarized as follows:
\begin{enumerate}

    \item In our pipeline, we propose a novel algorithm that enables us to complete contours that appear to be connected to each other in an illusory form.
    \item Our model is trained on just one single query image and does not need any training data.
    \item Our model does not need to know which regions of the image are masked or occluded, i.e., we remove the dependency of the algorithm on the guiding mask (a guiding mask is a mask that informs the model on where the missing regions are located at).    
    \item We also introduce two metrics to produce a stopping criterion to know when to stop training before the model fully overfits to the incomplete image, i.e., we guide the model to stop when the completed image is produced.
    
\end{enumerate}

\begin{figure}[!b]
  \centering
    \includegraphics[width = 0.45\linewidth]{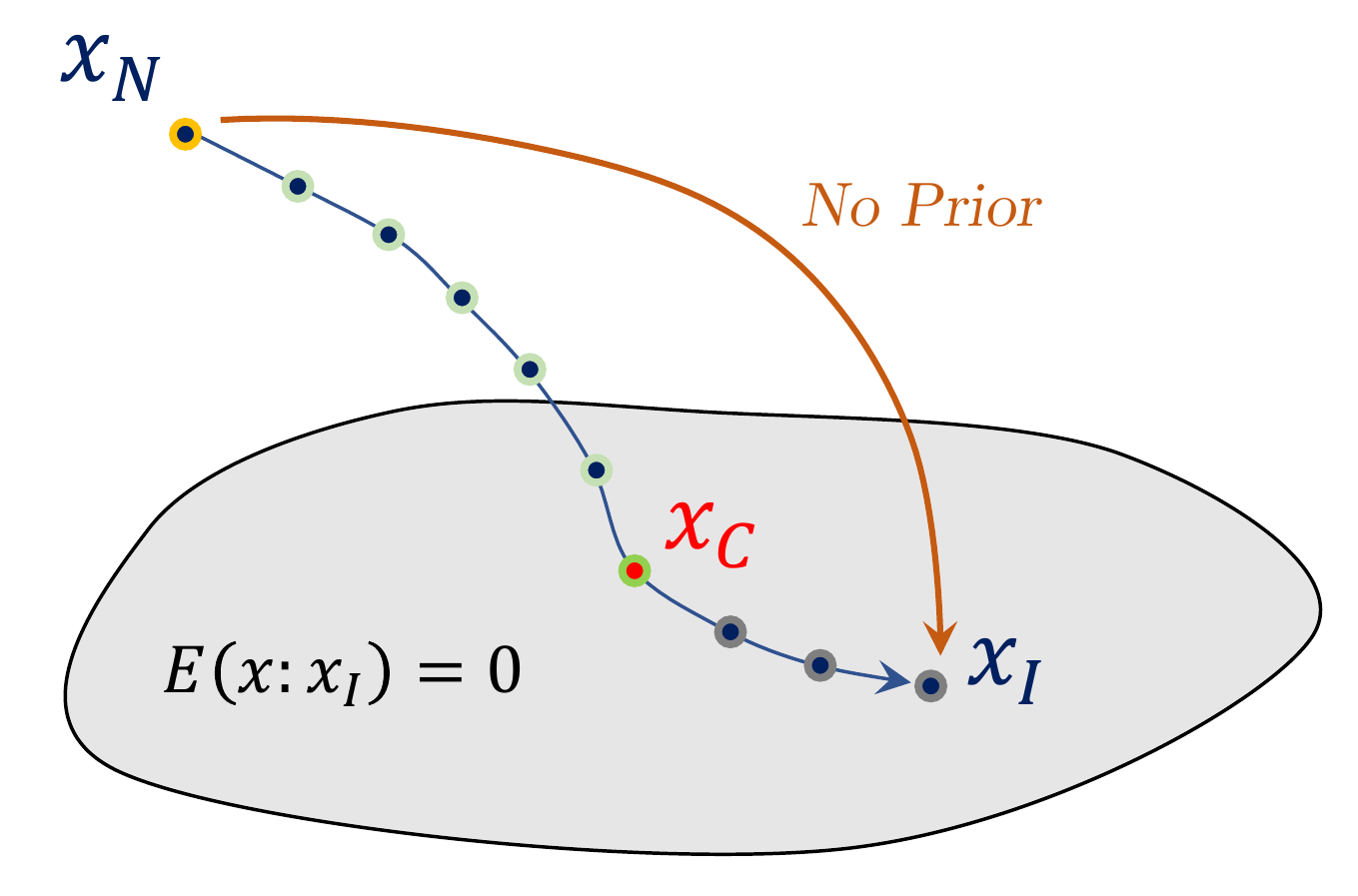}
    \caption{The trajectory from random noise $X_{N}$ to the incomplete image $X_{I}$ in image space. The network will pass a completed version of the image, $X_{C}$, throughout this trajectory.}
    \label{fig:figure_path}
\end{figure}

\section{Methods}
\label{methods}

Our eyes are trained to predict a missing region of an occluded object within a scene. We can easily perceive or make guesses about parts of objects or shapes that we do not necessarily see. Even when we are looking at an image, we might guess about the shape, property, or other attributes of an unknown object within a scene. Such capability extends beyond just known objects or shapes. We can look at a disconnected set of contours and guess what the connected form may look like. This capability is rooted in our prior knowledge about the world. (see Figure \ref{figure_kaniza}).

\begin{figure*}[!t]
	\centering
	\includegraphics[width = 0.95\textwidth]{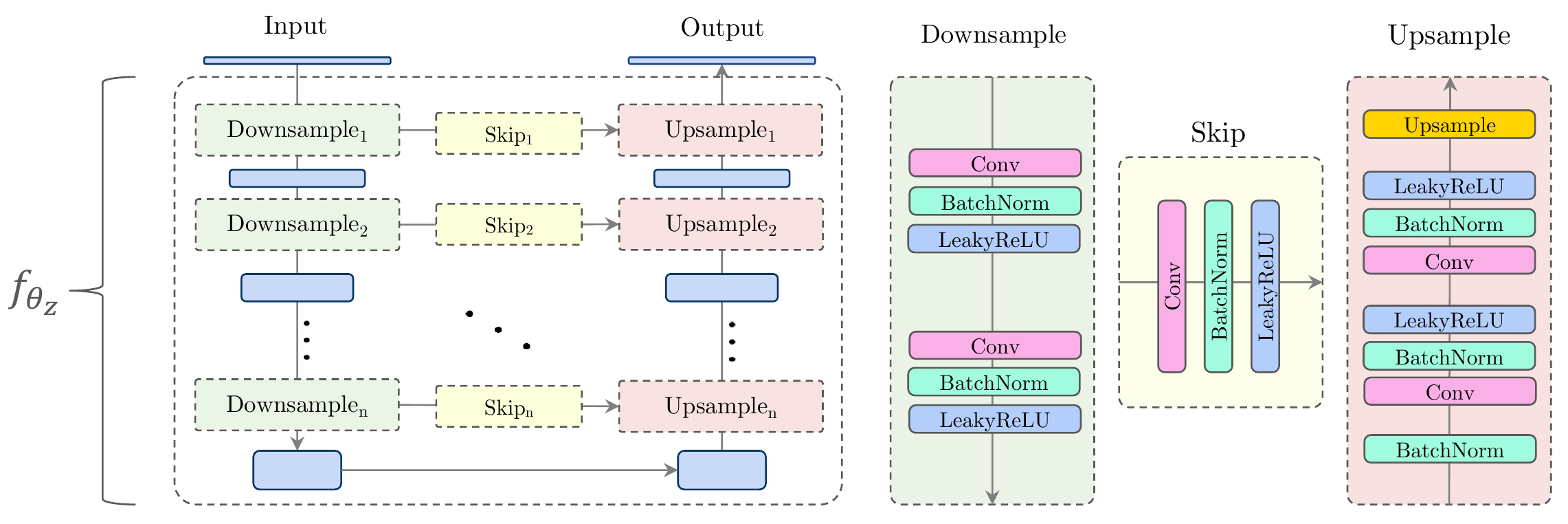}
    \caption{This figure shows the architecture of our model, which is based on the UNet-like hourglass architecture with skip-connection used in \cite{ulyanov2018deep}.}
    \label{figure3}
\end{figure*}

In this work, we aim to achieve a similar capability using deep generative networks. 
Most neural networks that we work with these days are trained with a massive amount of data and one might think that this is the only way that a neural network can obtain prior information. Authors of Deep Image Prior (DIP) \cite{ulyanov2018deep} suggest that the convolutional architecture can capture a fair amount of information about image distribution. They show that the hourglass architectures like {UNet \cite{ronneberger2015u} can show some good performances in some inverse problems such as image denoising, super-resolution, and inpainting.

\begin{figure}[!b]
  \centering
  \includegraphics[width = 0.5\linewidth]{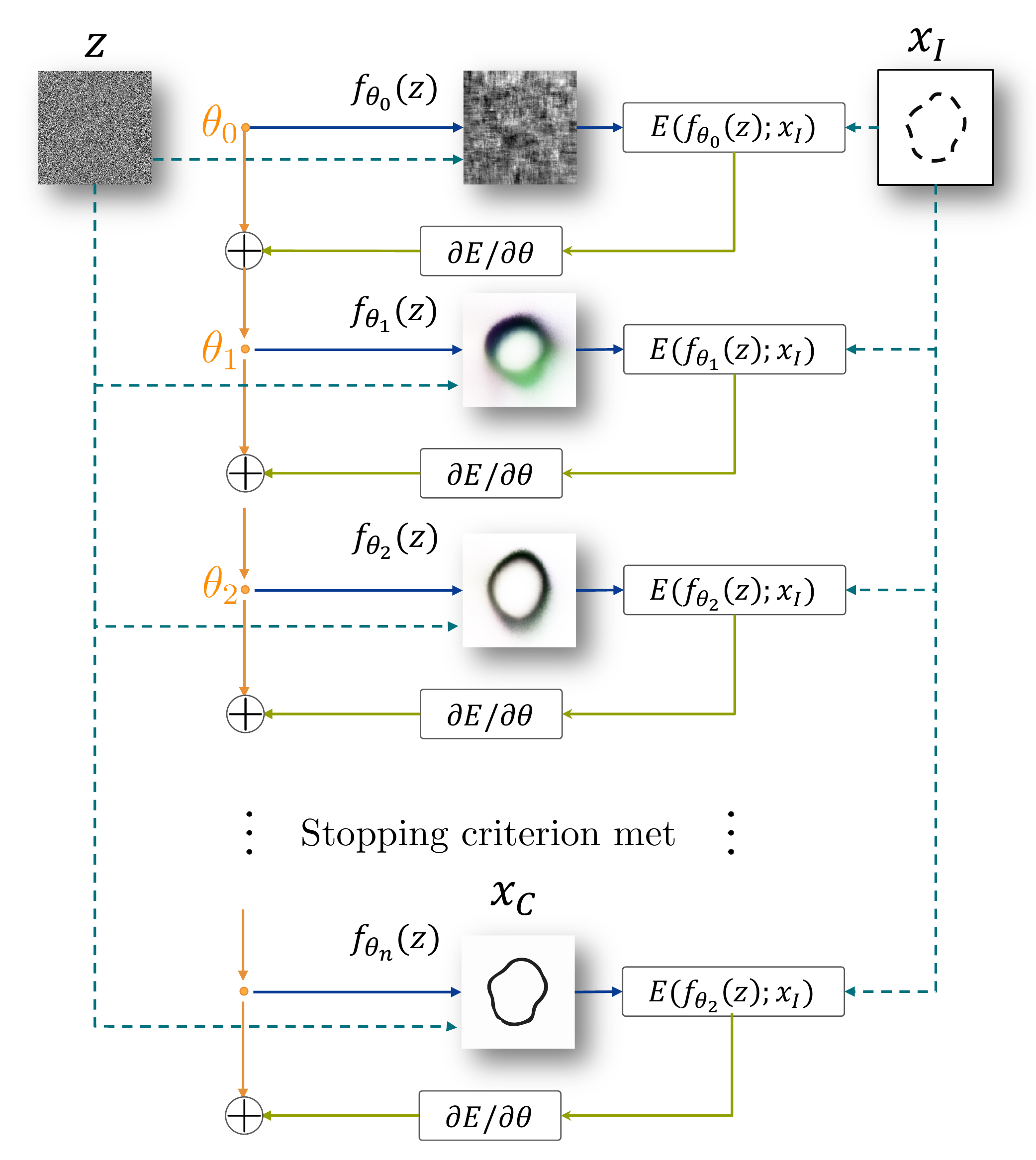}
  \caption{This figure shows our iterative process to complete the fragmented contours of an image given as input to our pipeline.}
  \label{fig:figure_iterative}
\end{figure}

In this work, we focus on completing fragmented contours end-to-end just by using a single image. To be able to address this problem, we first look at a similar problem in image editing, known as image inpainting. Image inpainting is the task of completing an image where some regions of that image are covered or filtered by a mask. In image inpainting, the generative model receives a masked image with the mask that guides the algorithm to fill in those missing regions. Although in the problem of contour completion, we have a very similar goal, the additional challenge that we suffer from is that we do not necessarily have a mask that covers the regions of interest for us. For example, when we look at Figure \ref{figure_kaniza} (left), we are not provided that which regions of the image are incomplete by a guiding mask. Our brain figures this out just by looking at the form and predicting those missing regions.

Inspired by the image inpainting work of DIP \cite{ulyanov2018deep}, we propose a novel algorithm for the contour completion problem (see Figure \ref{figure3}), where, unlike DIP, we do not have a guiding mask to know where to fill in the missing regions of our disconnected contours. 

Let us assume that we are given a degraded image $x_{I}$ containing a fragmented contour. We propose an iterative process (see Figure \ref{fig:figure_iterative}) that can connect those discontinuities and glue those fragmented pieces together as follows. We propose an hour-glass model structure ($f$) that is initially set up with completely random parameters ($\theta_0$) at first. Through an iterative process, we start feeding our network with a fixed random tensor noise $z$ signal and obtain the inferred output ($f(z)$) from that network. We then back-propagate the difference between the inferred output and the incomplete image to the network. We then repeat this process until the difference between the generated outcome of the network ($f_\theta(z)$) and the incomplete image ($x_I$) gets smaller and smaller and finally overfits the incomplete image ($x_I$). In this work, we propose a novel error metric to backpropagate in the model and update its weights. we set the metric in a way that enables us to complete the incomplete image before it overfits the incomplete image. This is where the magic of our algorithm happens. We also propose a stopping criterion, so that when the image is complete, we no longer overfit the outcome of the model and instead produce a plausible connected set of fragmented contour pieces. As illustrated in Figure \ref{fig:figure_path}, this trajectory will pass through a complete version of the image in image space, which is close to the actual connected ground truth $x_{gt}$, which we do not have access to directly.

\subsection{Energy Function}

We can model this iterative process mathematically by maximizing a posterior distribution. Let us assume that the optimal image $x^*$ that we want to achieve is on a path that connects a random tensor noise $z$ to the incomplete image $x_I$. With this assumption, we can eventually overfit any random tensor noise to the incomplete image $x_{I}$, and we can formulate the posterior distribution of our wanted optimal image $x^*$ as follows:
\begin{equation*}
x^{*} = \argmax_x p(x|x_I),
\end{equation*}

To better recapitulate what we want to achieve using our generative model, we solve an energy minimization problem on the parameter space of the model, rather than explicitly working with probability distributions and optimizing on $x$ (image space). Thus, we solve an energy minimization problem that incorporates the incomplete image ($x_I$) and model parameters ($f_\theta(z)$):

\begin{equation*}
\theta^{*} = \argmin_\theta E(f_{\theta}(z);x_{I}),\qquad x^{*}=f_{\theta^{*}}(z)
\end{equation*}

As shown in Figure \ref{fig:figure_iterative}, the pipeline starts from a random initialized set of parameters $\theta$ and updates those weights until it reaches a local minimum $\theta^{*}$. The only information provided for the network is the incomplete image $x_I$. 
When we reach the optimal $\theta^{*}$, the completed image is obtained as $x^{*} = f_{\theta^{*}}(z)$ where $z$ is random tensor noise. In this work, we use a U-Net architecture with skip connections as the generator model.

As we mentioned previously, in this work we were inspired by an earlier work known as Deep Image Prior (DIP) \cite{ulyanov2018deep}. In this work, the authors suggested a mean-squared-error loss term  that enables the network to compare the output of the generator to the incomplete input image: 

\begin{equation}
E(x;x_{I})=\left \|(x-x_{I})\odot m \right \|\ ^{2}
\label{eq:base}
\end{equation}

where $ x_{I} $ is the incomplete image with missing pixels in correspondence of a binary mask $m \in \left \{ 0,1 \right \}^{H \times W}$ and $\odot$ operator is for point-wise multiplication of two image matrices. In the inpainting tasks, the existence of a mask is essential as the algorithm needs to know where to fill in the missing area, whereas, in our work, we wanted to know whether the network can perform completion on its own without the need for the mask. In other words, is it possible for the network to predict where to fill in at the same time that it is trying to reconstruct the incomplete image through the iterative process? To answer this question, we tried to solve a much harder problem in which the mask is not provided to the model and the model is agnostic to it. To better understand how a solution could be hypothesized for this problem, we first imagine that we want to consider all the available regions in our image that could be potential places to fill in, i.e., we set the mask in the previous formula \ref{eq:base} to be equal to the incomplete image $x_I$. This is problematic as the model quickly tries to fill in all white space and quickly reconstructs the incomplete image by doing so. On the other hand, we can take the inverse problem of the current problem, where the model tries to just fill in the regions that fragmented contour lives in. Taking these two at the same time, we came up with a novel loss term for energy minimization term that helps us remove the need for the mask in the case of the contour completion problem:

\begin{multline} \label{eq:2}
	E(x;x_{I}) = \alpha \times \overbrace{\|(x-x_{I})\odot x_{I}  \|\ ^{2}}^{\text{missing region reconstruction loss}}
	+ (1-\alpha) \times \overbrace{\left \| (1-x)\odot(1-x_{I}) - (1-x_{I})\odot(1-x_{I}) \right \|\ ^{2}}^{\text{original contour reconstruction regularization term} }
\end{multline}
In this term, we introduce a linear combination of the two loss terms, where one focuses on reconstructing the missing regions in the foreground, and one focuses on avoiding inpainting regions in the background. The logic behind this is that, if we assume the original image to be representative of the mask, then the model tries to reconstruct in all white regions (the foreground), and in the inverse problem we just want to reconstruct the regions that are already part of the ground truth. 

\subsection{Stopping Criteria}

As shown in Figure \ref{fig:figure_path}, knowing when to stop iterating to the over-fitted model is a key to obtaining a completed shape. Therefore, we equipped our model with a criterion that uses two individual novel terms to know when to stop and output the result of the network. These two metrics expand the capability of the generator network beyond what it does currently and achieve a full end-to-end contour completion model that trains and infers on a single image of divided contour fragments.  These new terms are:  \textit{reconstruction\_score} ($\rho$) and \textit{overfit\_score} ($\omega$). 

\subsubsection{Reconstruction Score}
The first score that this paper suggests is the reconstruction score, i. e., we have to make sure that the model is trained enough that it can reconstruct at least the entire set of fragmented contours within the image. This is a trivial score and to compute the \textit{reconstruction\_score} ($\rho$), we apply a $k$-dimensional tree (KDTree) nearest-neighbor lookup to find the ratio of points in the original incomplete image ($x_0$). This score ranges from $[0 - 100]$.

\subsubsection{Overfit Score}
It is evident that the model overfits the fragmented contours. This is due to the fact that the error in our loss term is minimized as the $x$ overfits to $x_I$, i. e., replacing $x$ with $x_I$ in the loss term would give us zero. As we hypothesize iterative process also produces the complete image before it overfits to the incomplete image, we can imagine that at some point the image is complete ($x_C$) and does not need to be fine-tuned any more to overfit to $x_I$. We suggest a new score called \textit{overfit\_score}. \textit{overfit\_score} determines how much of the reconstructed outcome is over the number of pixels that are already in the incomplete image ($x_I$). To compute the \textit{overfit\_score} ($\omega$), we apply a $k$-dimensional tree (KDTree) nearest-neighbor lookup of points in the outcome of the input image and see what portions of those points are novel and not already in the incomplete image ($x_I$). Similar to \textit{reconstruction\_score}, the \textit{overfit\_score} also ranges from $[0-100]$.

\subsubsection{Combined Score}
To be able to find the best possible set of completed contours, we combine the two and have a loop that tries to achieve close to full reconstruction and avoids over-fitting at the same time. This is what we call an "ideal" stopping point in the contour completion problem. In each run throughout all iterations, we pick an output of the network that minimizes a dissimilarity term:

\begin{equation}
\delta = \sqrt{(\rho- 100)^{2} + (\omega - \gamma)^{2}}
\end{equation}

where $\delta$ represents our dissimilarity score. The \textit{reconstruction\_score} and \textit{overfit\_score} are obtainable given network output and the incomplete image. Ideally, we want to achieve an output image that has a \textit{reconstruction\_score} equal to 100 and an \textit{overfit\_score} of $\gamma$ which is a hyperparameter that is dependent on the image and complexity of the shape. Empirically, we observed that this value is highly correlated with the distance between gaps that are observed in fragmented contours, i. e., the larger the gap results in a larger $\gamma$ value. We will discuss this in more detail in the next section (see Section \ref{exp}).

\begin{figure}[t]
  \centering
    \includegraphics[width = 0.35\linewidth]{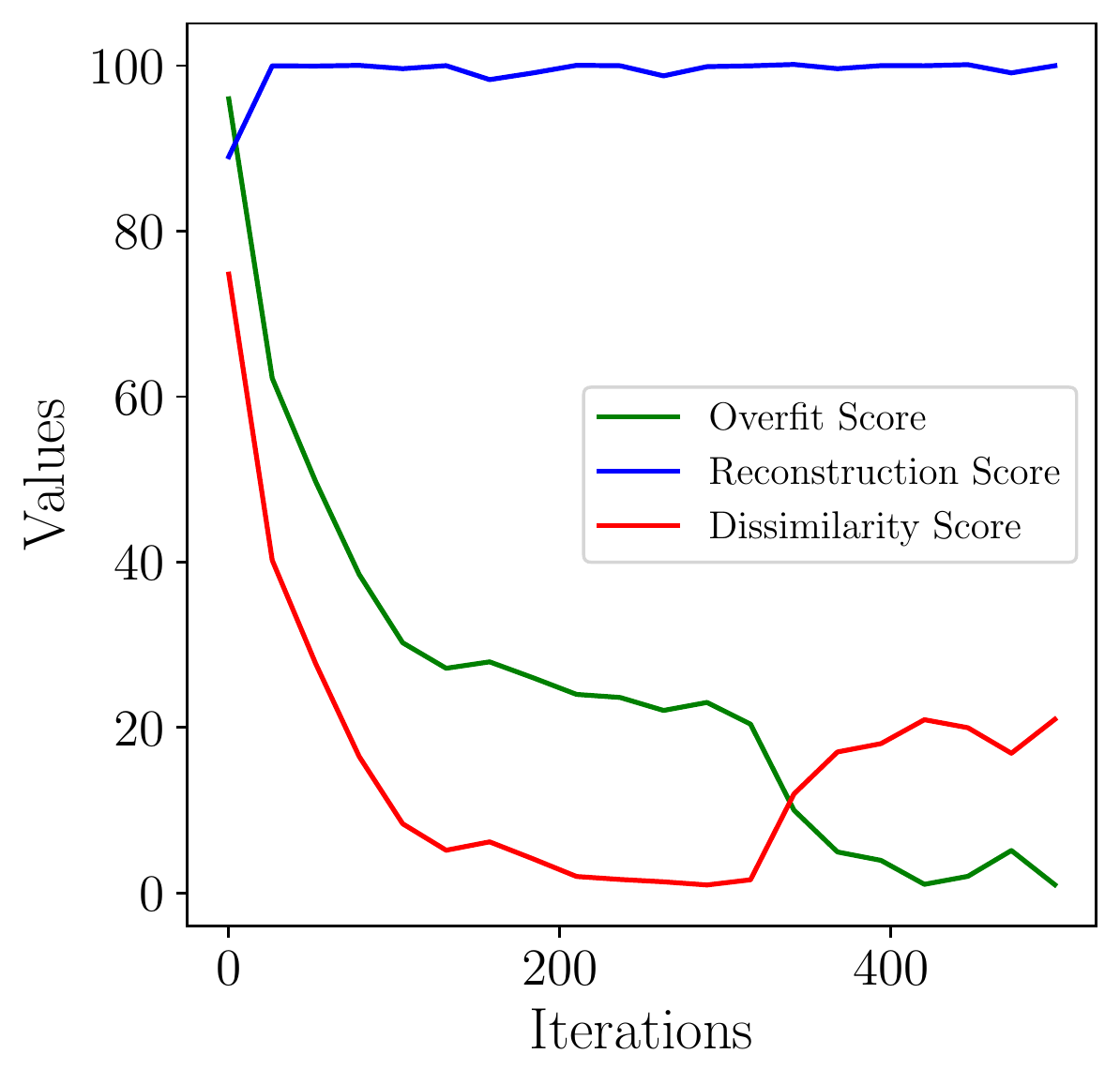}
    \caption{
    This example shows how different scores change throughout a single run. All three scores change in the range of $[0,100]$. Our goal is to maximize \textit{reconstruction\_score} and minimize the \textit{overfit\_score}, but we should consider that the minimization lower bound is data dependent and is not zero.}
    \label{figure6}\vspace{-4mm}
\end{figure}

For one sample image, we computed the two metrics \textit{reconstruction\_score} and \textit{overfit\_score} and the combined value of dissimilarity ($\delta$) and showed how these values change (see Figure \ref{figure6}). Our initial observations show that the \textit{reconstruction\_score} will increase to 100 quickly for the incomplete image indicating that the already existing fragments of the contours have been reconstructed in the output. 

However, as mentioned previously, we cannot solely rely on this score since we also want to minimize the overfitting. Remember that our goal is to produce an output that: a) preserves the original contours in the incomplete image and b) fills in the gaps between the fragmented contours.
It is evident that \textit{overfit\_score} decreases throughout an iterative run of our process until it reaches zero. The \textit{dissimilarity} will also decrease along with the overfit to a point, then it will increase, as the model tries to reproduce the incomplete image. This is where an ideal $\gamma$ value can be picked, i.e., where to stop when the reconstruction is good but we have not done a full overfit to the incomplete image.
Thus, one should pick the value of $\gamma$ empirically in the scenario that the ground truth is not available, whereas, assuming that the ground truth is available, we can easily compute the best $\gamma$ value. 
In our experiments, we tried two datasets of images with different gap sizes. We observed that the best the $\gamma$ for one set of samples is $\sim 5$ (the set with shorter gaps) while it is $\sim 23$ for samples from the other set, i. e, the set with longer gaps (see Figure \ref{fig:examples} for some completed examples).

\begin{figure*}[!b]
	\centering
	\begin{tabular}{c@{\hskip 1pt}c@{\hskip 1pt}c@{\hskip 1pt}c@{\hskip 1pt}c@{\hskip 1pt}c@{\hskip 1pt}c@{\hskip 1pt}c}
		\fbox{\includegraphics[width = 0.107\textwidth]{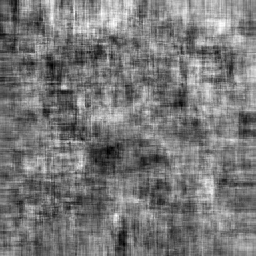}} & \fbox{\includegraphics[width = 0.107\textwidth]{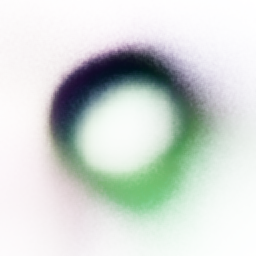}} & \fbox{\includegraphics[width = 0.107\textwidth]{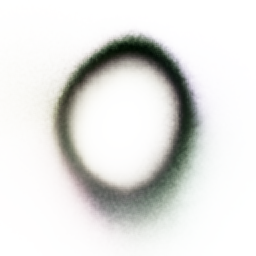}} & \fbox{\includegraphics[width = 0.107\textwidth]{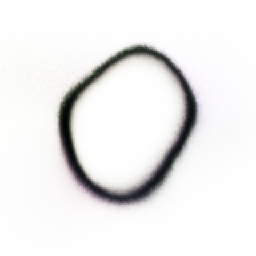}} & \fbox{\includegraphics[width = 0.107\textwidth]{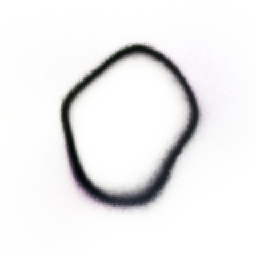}} & \fbox{\includegraphics[width = 0.107\textwidth]{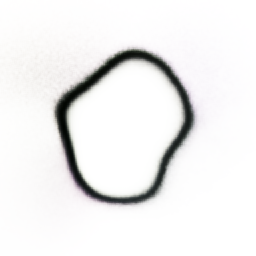}} & \fbox{\includegraphics[width = 0.107\textwidth]{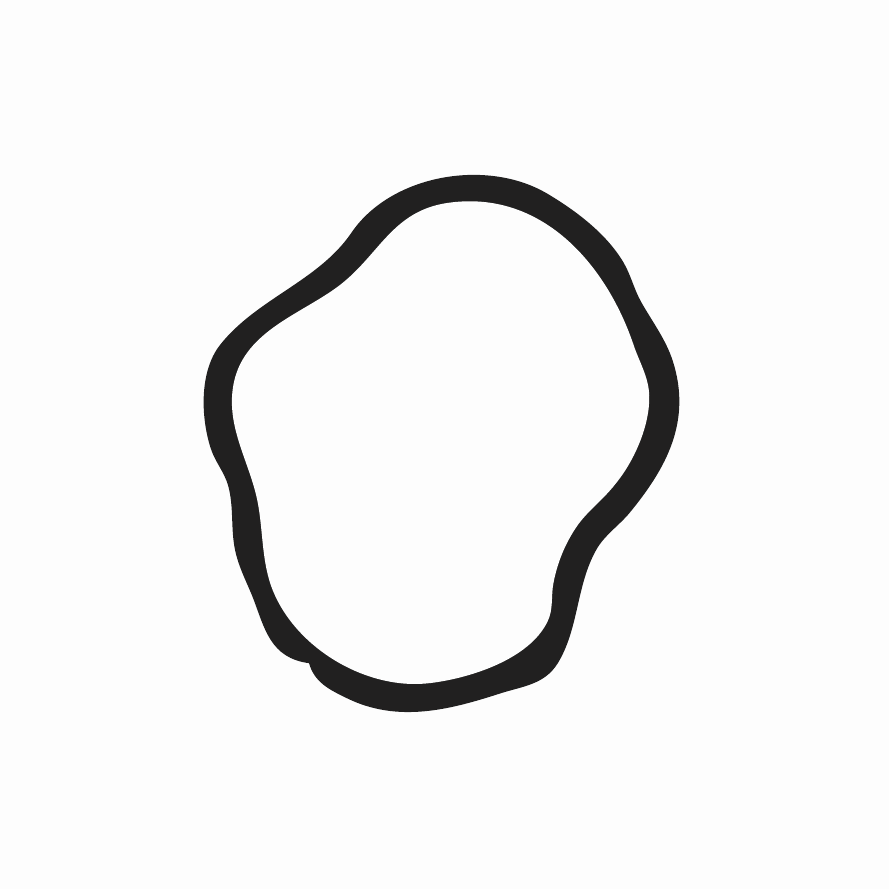}} & \fbox{\includegraphics[width = 0.107\textwidth]{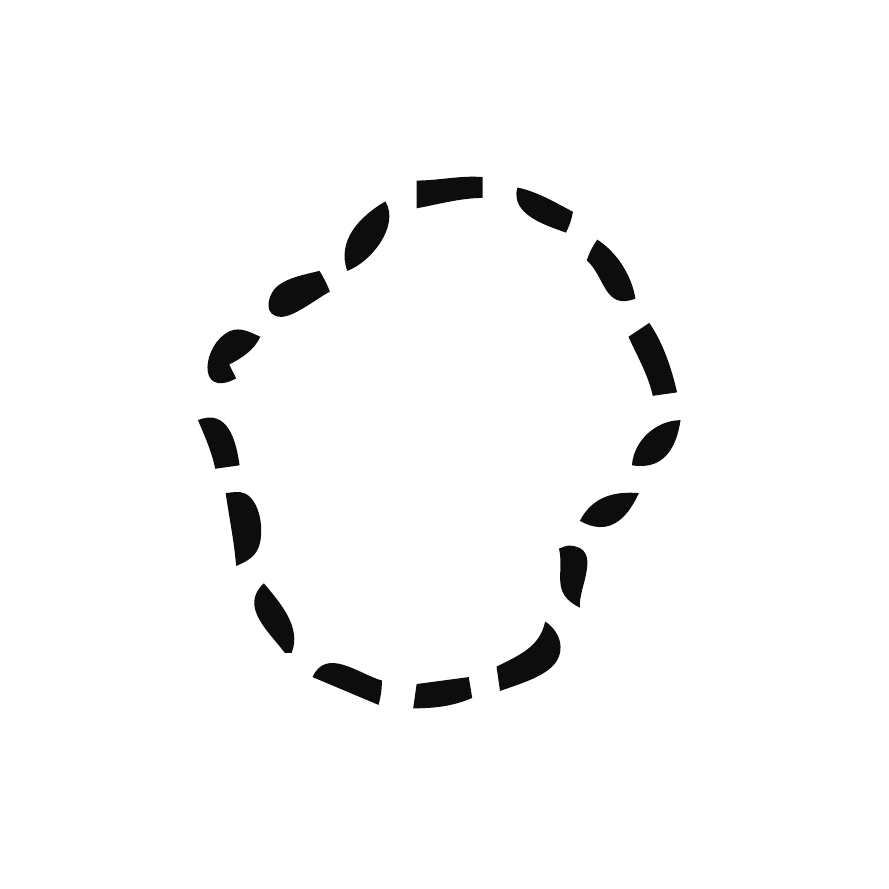}} \\
		\fbox{\includegraphics[width = 0.107\textwidth]{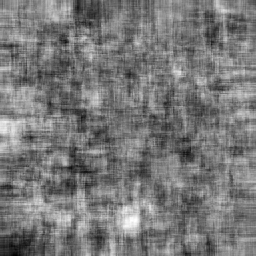}}   & \fbox{\includegraphics[width = 0.107\textwidth]{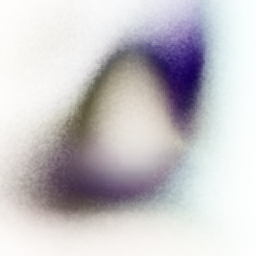}}   & \fbox{\includegraphics[width = 0.107\textwidth]{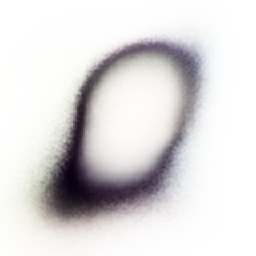}}   & \fbox{\includegraphics[width = 0.107\textwidth]{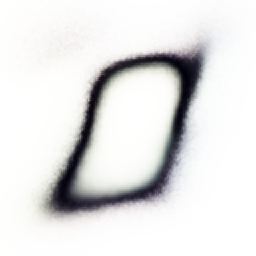}}   & \fbox{\includegraphics[width = 0.107\textwidth]{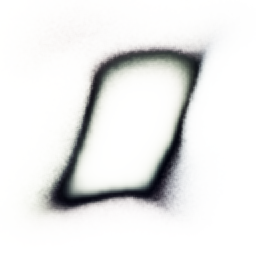}}   & \fbox{\includegraphics[width = 0.107\textwidth]{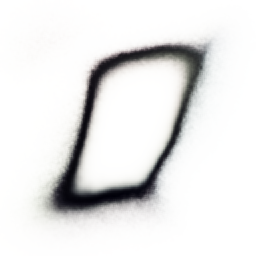}}   & \fbox{\includegraphics[width = 0.107\textwidth]{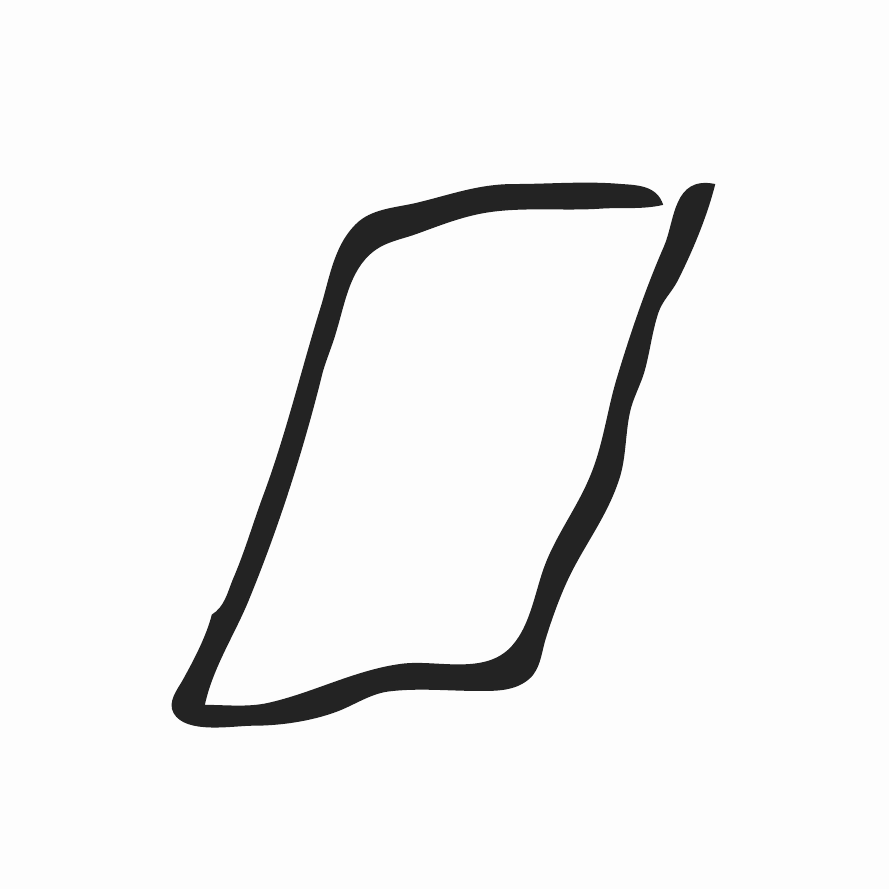}}   & \fbox{\includegraphics[width = 0.107\textwidth]{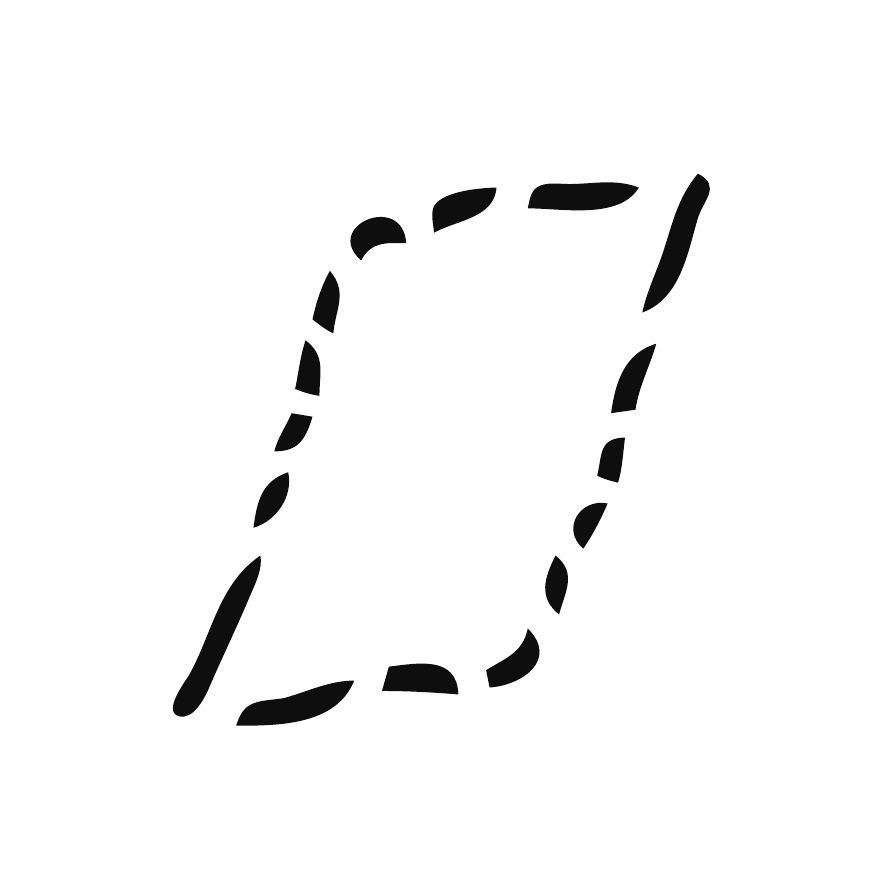}}   \\
		\fbox{\includegraphics[width = 0.107\textwidth]{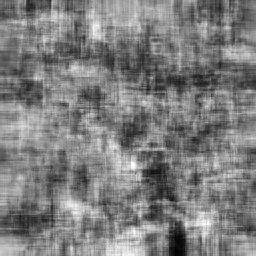}}    & \fbox{\includegraphics[width = 0.107\textwidth]{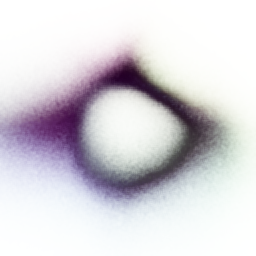}}    & \fbox{\includegraphics[width = 0.107\textwidth]{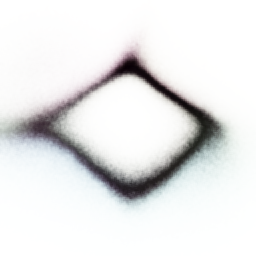}}    & \fbox{\includegraphics[width = 0.107\textwidth]{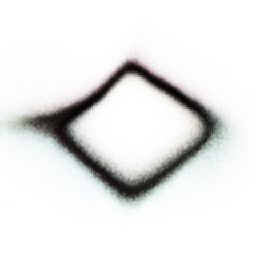}}    & \fbox{\includegraphics[width = 0.107\textwidth]{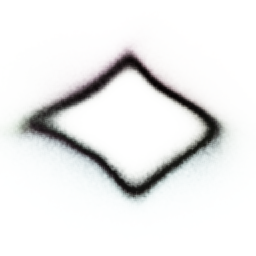}}    & \fbox{\includegraphics[width = 0.107\textwidth]{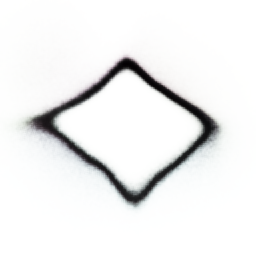}}   & \fbox{\includegraphics[width = 0.107\textwidth]{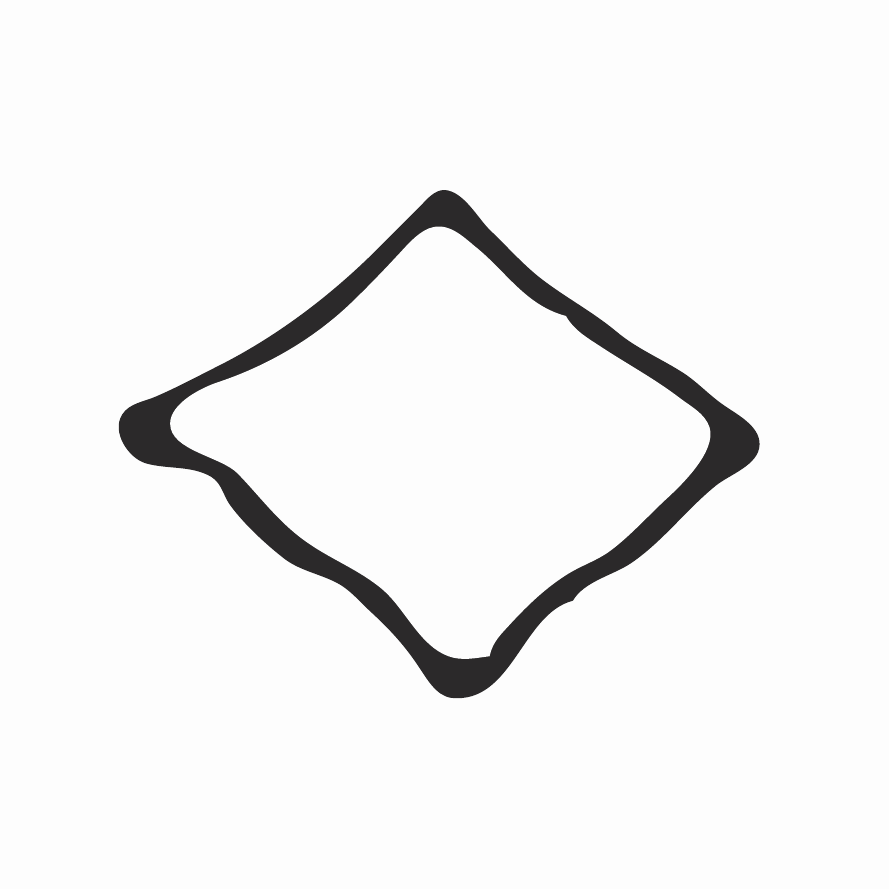}}    & \fbox{\includegraphics[width = 0.107\textwidth]{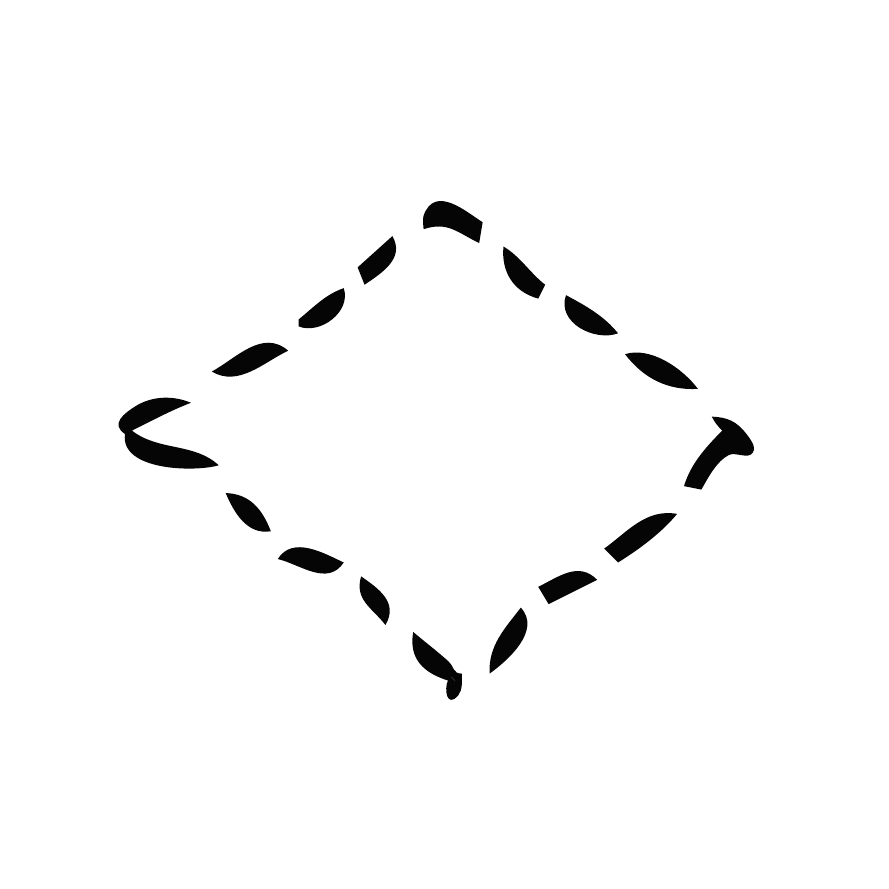}}                     
	\end{tabular}
	\caption{Evolutionary process of the deep structure prior. The right column shows the incomplete shapes given to the model and the rest of the columns show how the model is overfitting gradually to produce the incomplete shapes. In each column, we are showing an intermediate iteration of this process. The loss-term setup enables our pipeline to let the completed image appears during this iterative process. }
	\label{fig:examples}
\end{figure*}
\section{Experiments and Results}
\label{exp}

Performing unsupervised contour completion is a difficult task to benchmark as one can never know what fragments exactly are connected to each other in a real-world scenario. This makes the problem of contour completion a hard problem to solve. In this paper, we tried to create artificial shapes that are occluded by some masks and then tried to see if our model can regenerate the missing pieces and glue those divided contours together. To demonstrate our model's behavior, we will conduct experiments on datasets created for this task and will report on them in this section. To compare network results in different settings, we will use pixel-wise Mean Squared Error (MSE) and Intersection over Union (IoU) between the produced result of the network and unmasked ground truth data and the reconstructed image on black pixels (where contours live).

\subsection{Data}
We prepared two datasets, one labeled ``Simple" and one ``Complex", in accordance with the number of gaps in each shape. Both datasets contain nine different categories of shapes. In order to generate the \textit{Complex dataset}, we used FlatShapeNet \cite{reevald2022} which is a dataset for the educational game Ariga. The dataset includes the following categories: Circle, Kite, Parallelogram, Rectangle, Rhombus, Square, Trapezoid, Triangle and Overlap. The ``overlap" category contains images that are made as a mixture of two shapes that are overlapping from the previous categories. These are some standard shapes with a few gaps in  \textit{simple dataset}, while the \textit{complex dataset} has some hand-drawn shapes with fragmented lines and more gaps that produce more variety in general. For each instance, a ground truth image is available for comparison. 
Most of our experiments have been conducted using the \textit{complex dataset} in order to evaluate the generalization of our approach. For the analysis of how $\gamma$ values should be set for each shape, we used the \textit{simple dataset} as a reference.

\subsection{Evaluation}
In this section, we compare our model to the original Deep Image Prior (DIP) \cite{ulyanov2018deep} inpainting model. DIP's inpainting module accepts a degraded image and a binary mask corresponding to that image. In order to make a fair comparison, instead of providing a binary mask, we used the incomplete images both as input and as a mask in order to see whether it can produce a result similar to ours. 

For DIP, we run the iterative process for a maximum number of 2500 iterations with the U-net backbone.
We used the exact same architecture and setting in our model for a fair comparison. Using our ground truth dataset images, we calculate the MSE loss between the network output and ground truth during each iteration instead of relying on our stopping mechanism described in the previous section. We then store the output with minimal loss throughout all the iterations. Finally, we select the best output among all iterations, report the MSE and IoU with the ground truth, and save the iteration number which resulted in the lowest MSE. Table \ref{table1} compares the results that are obtained using the DIP method, the DSP method (ours), and the difference between raw images and the ground truth. We have presented the average MSE-loss, average IoU, and the average number of iterations for the best output for different methods. As can be seen from the table, our model improves both MSE and IoU between the incomplete image and ground truth in fewer iterations. The DIP method can neither generate a better result than the raw image nor provide stopping criteria to prevent overfitting.
We provide a more detailed analysis of this result in Figure \ref{figure7}. As results show, our algorithm not only provides a much faster convergence but also consistently provides a better-completed image (consistently less MSE loss and better IoU), whereas it is challenging for the DIP method to accomplish better results without a guiding mask.  

\begin{figure*}
    \centering
    \begin{tabular}{c@{\hskip 0pt}c@{\hskip 0pt}c}
        \includegraphics[width=0.33\textwidth]{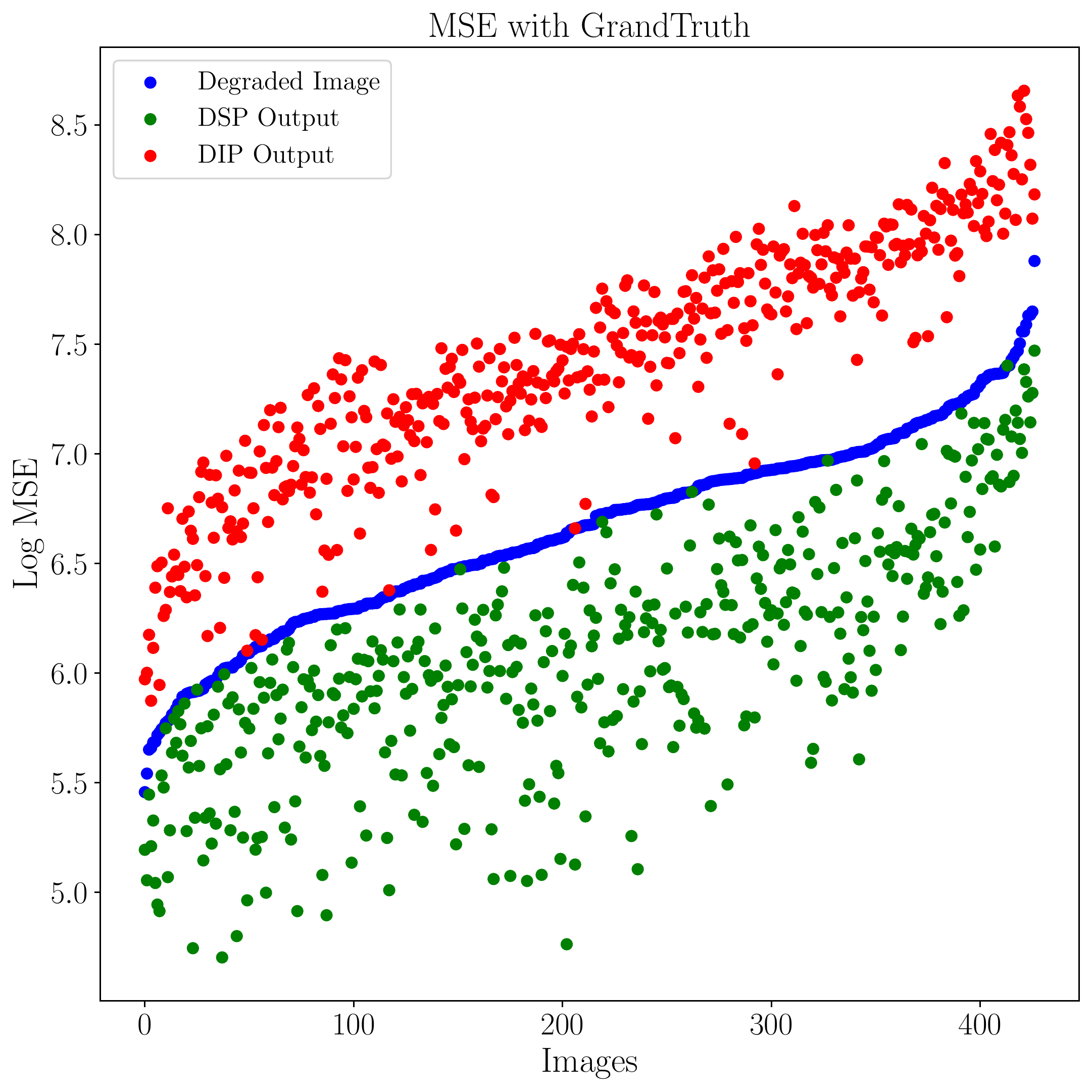} &  \includegraphics[width=0.33\textwidth]{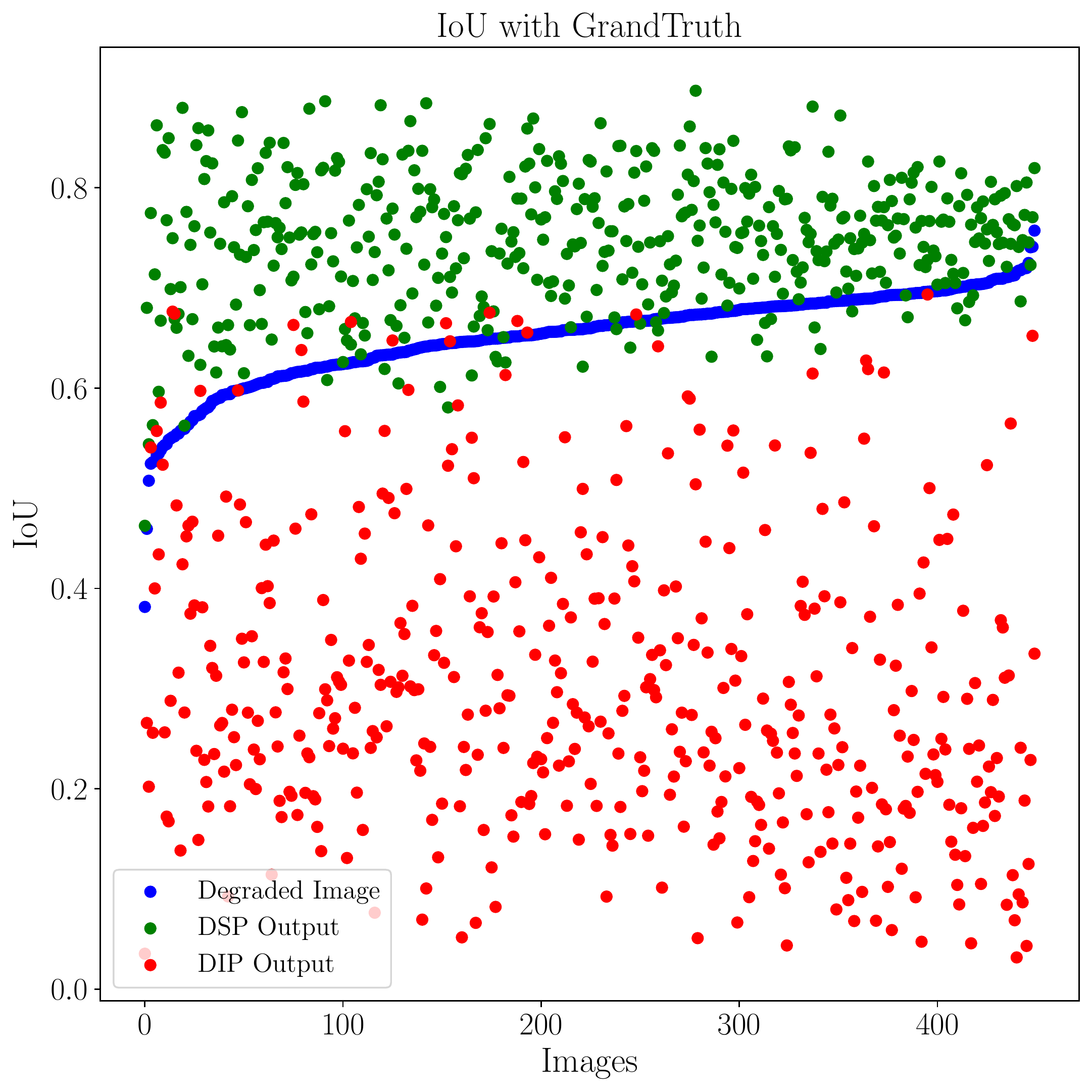} &
        \includegraphics[width=0.33\textwidth]{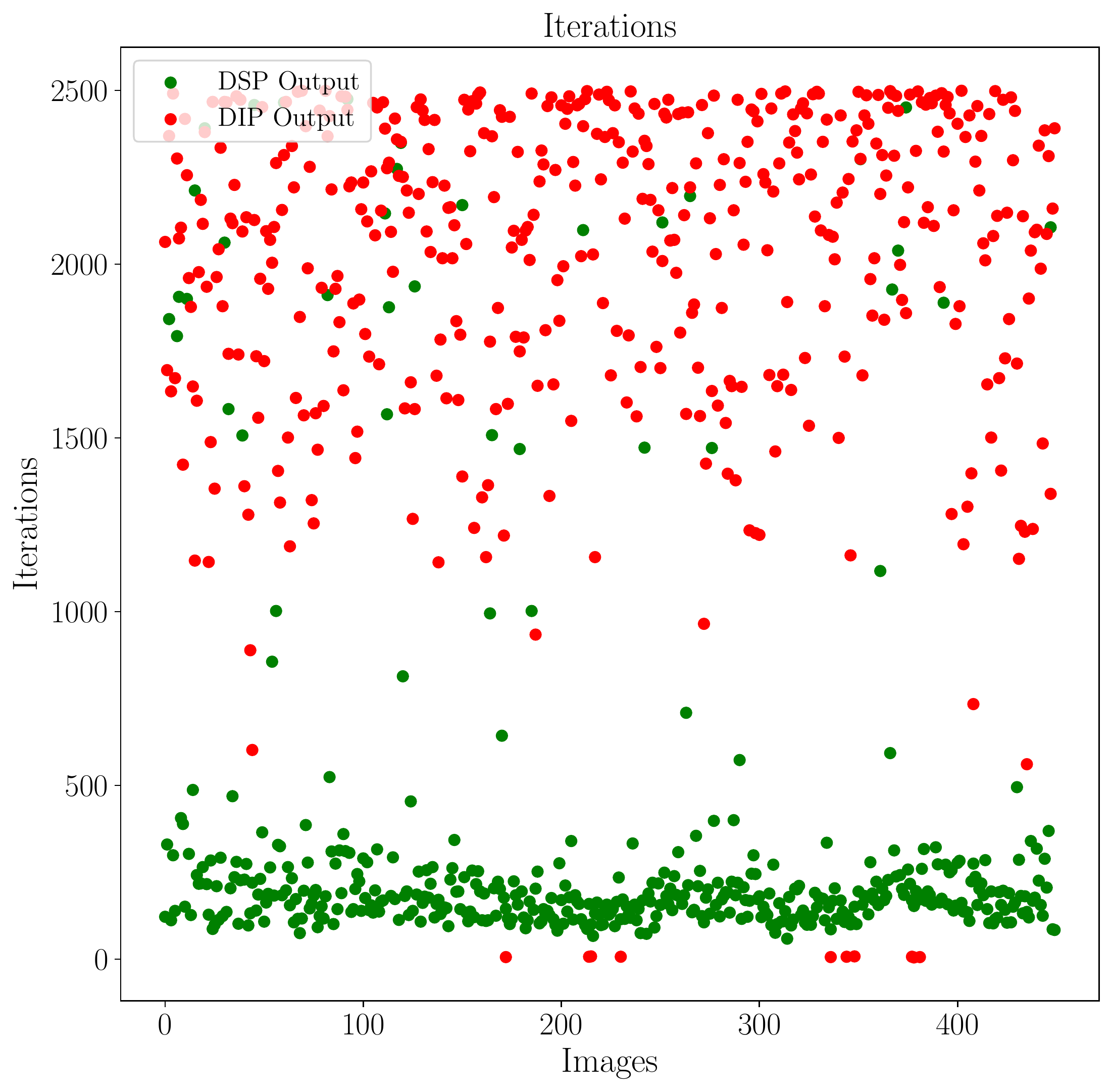}\\
        \textbf{(a)} & \textbf{(b)} & \textbf{(c)} 
    \end{tabular}
    \caption{In this Figure, we compare the completed images from our algorithm, DSP, (shown in green) with the best obtained completed images from the DIP method (shown in red). We compare MSE loss between the degraded raw images that these algorithms started with (shown in blue) \textbf{(a)}  Mean Squared Error Loss: we clearly see that for almost all images, DSP (green) achieves a lower MSE than the incomplete images (blue) whereas, the DIP completed images either do not improve the MSE or even worsen that for the incomplete images. Note that, the MSE is computed to an available ground truth image hidden from our methods (the lower is better). \textbf{(b)} Intersection Over Union: here, we are looking at the IoU metric that specifies the amount of intersection over the union between the obtained images and the ground truth data. Again, we see that DSP produces images that are much closer to the ground truth (in most cases) whereas the DIP can not achieve a similar result. While we see few DIP completed images produce a better IoU than the degraded images (in terms of IoU), most of them are worse than the starting image (the higher is better).  \textbf{(c)} The number of iterations that are needed for each algorithm to obtain its best results. Here, we see that DSP can quickly produce the best outcome with the least MSE loss whereas the DIP algorithm's best results are when we run the iterative process for more iterations (the lower is better).  }
    \label{figure7}
\end{figure*}

\begin{table}[!t]
\caption{Average MSE and IoU values between the incomplete (Raw) images, the output of DIP and DSP methods, and ground truth for each image are provided in this table.}
\label{table1}
\begin{center}
\begin{small}
\begin{sc}
\begin{tabular}{lcccc}
\toprule
Method & MSE & IoU & Iterations & Time (s)\\
\midrule
Raw & 829.01 & 0.65 & 0 & 0\\
DIP & 2027.52 & 0.29 & 2013 & 54.68\\
DSP (Ours) & \textbf{490.84} & \textbf{0.74} & \textbf{359} & \textbf{9.55}\\
\bottomrule
\end{tabular}
\end{sc}
\end{small}
\end{center}
\vskip -0.1in
\end{table}

\subsection{Correlation of $\gamma$ with the Gap Size}
To better understand the $\gamma$ parameter of our combined score, we conducted the following experiment. A total of 12000 samples were obtained by merging all of our images from the two datasets, \textit{simple} and \textit{complex}. As we have access to the ground truth for each degraded image in the combined dataset, we can easily calculate \textit{reconstruction\_score} and \textit{overfit\_score} for each degraded-ground truth pair. As expected, we obtain a \textit{reconstruction\_score} of 100 for all samples, but the \textit{overfit\_score} varies among them. 
Intuitively, we hypothesized that an optimal value of overfit score should be intertwined with the total area of gaps. To test this hypothesis, we did the following experiment. We first define a function $\phi(x)$ which takes a binary, black and white image $x$ and returns the number of black pixels in it. Then we define a \textit{gap} term as follows:

\begin{equation}
gap = \frac{\phi(x_{gt})  - \phi(x_{I})}{\phi(x_{gt})}
\end{equation}

where $x_{I}$ is the incomplete image and $x_{gt}$ is the ground truth. In this case, \textit{gap} indicates the total area of the gap with respect to the whole shape. We found out that this term and the best result have a correlation of $\mathbf{97.43\%}$. This indicates that the value of $\gamma$ is highly correlated with the gap size, that is something expected in a way.

\subsection{Effect of $\alpha$}
\label{further}
We conducted additional experiments concerning how $\alpha$ affects the quality of reconstruction. In the previous section, we defined Equation \ref{eq:2} as the loss term that guides our iterative process. The term $\alpha$ specifies the amount of emphasis the model should place on reconstructing missing regions, rather than filling in fragmented contours. A lower $\alpha$ indicates a better grouping quality, as shown in Equation \ref{table2}. However, we will not achieve completion if we remove the first term completely from the loss by setting $\alpha=0$. Therefore, the first term should be kept, but its weight should be very low in order to achieve a good completion. On the other hand, if we set $\alpha=1$ and omit the second term, we lose the contour completion regularization term and obtain the same output as a vanilla deep image prior, which does not complete shapes.
\begin{table}[!h]
\caption{
For this experiment, we ran the model over a subset of \textit{complex} dataset with 500 incomplete images at various levels of alpha for 250 iterations. After the image completion is done, we compared the evaluation metrics between the completed image and the ground truth to examine the performance of the model for different values of alpha.\\}
\centering
\begin{tabular}{c c c c c} 
 \toprule
 Alpha & MSE & IoU & Iterations & Time (s)\\ [0.25ex] 
 \hline\hline
 0    & 17240.44 & 0.04 & 0 & 0\\ 
 0.05 & 619.82   & 0.71 & 221 & 7.75\\ 
 0.1  &  456.77 & \textbf{0.76} & 210 & 7.35 \\ 
 0.15 & \textbf{456.52} & 0.75 & 184 & 6.37 \\ 
 0.2  & 485.53   & 0.74 & 157 & 5.44 \\ 
 0.25 & 527.35   & 0.73 & 151 & 5.25 \\ 
 0.3  & 551.23   & 0.72 & 151 & 5.25 \\ 
 0.35 & 602.57   & 0.71 & 155 & 5.31 \\ 
 0.4  & 644.36   & 0.69 & 160 & 5.34 \\ 
 0.45 & 682.33   & 0.68 & 172 & 5.77 \\ 
 0.5  & 749.29   & 0.66 & 183 & 6.13 \\ 
 0.55 & 825.89   & 0.63 & 197 & 6.80 \\ 
 0.6  & 906.45   & 0.60 & 206 & 7.10 \\ 
 0.65 & 996.34   & 0.57 & 222 & 7.70 \\ 
 0.7  & 1132.19  & 0.52 & 228 & 7.91 \\ 
 0.75 & 1294.25  & 0.48 & 228 & 7.92 \\ 
 0.8  & 1546.86  & 0.43 & 225 & 7.78 \\ 
 0.85 & 2111.38  & 0.36 & 216 & 7.40 \\ 
 0.9  & 4031.41  & 0.29 & 185 & 6.32 \\ 
 0.95 & 7772.83  & 0.16 & 107 & 3.70 \\ 
 1    & 12258.55 & 0.06 & 11  & 0.36 \\ 
\hline
Raw & 843.23 & 0.65 & 0 & 0\\ 
 \bottomrule
\end{tabular}
\label{table2}
\end{table}

\subsection{Effect of Receptive Field Size}
To better understand the effect of receptive field size on our algorithm, we test the following hypothesis: can models with bigger receptive field size complete shapes with bigger gaps? In Table \ref{tab:receptive_field}, we report showing the results of this experiment. As we can see, the bigger the receptive field size, the more complete shapes we can reconstruct using DSP. 
{\renewcommand{\arraystretch}{1.05}\begin{table}[!h]
\caption{In this table, we show the effect of the receptive filter size on our algorithm's capability to fill in bigger gap sizes. The numbers in this table are showing the percentage of the time that DIP was successful to complete shapes with each gap size and corresponding receptive field size. As predicted, the bigger the filter size, the more successful the algorithm is in filling in the gaps. \vspace{1mm} }
    \centering    
    \begin{tabular}{c|cccc}
    \toprule
      Receptive  & \multicolumn{4}{c}{Gap Size}\\
       \cline{2-5}
       Filter size  & 8 & 10 & 12 & 15  \\\hline\hline
     3 & 66.66\%   &  56.0\% & 26.13 \%  & 20.26\%\\\hline
      5 & 85.33\% & 66.66\% & 54.13\% & 26.13\%\\\hline
      7 & \textbf{85.33}\% &  \textbf{76.0} \% & \textbf{54.66}\% & \textbf{26.93}\%\\
    \bottomrule
    \end{tabular}
    \vskip -2.5mm
    \label{tab:receptive_field}
\end{table}}
\section{Implementation Details}
As shown in Figure \ref{figure3}, we use a model with 5 layers and 128 channels for downsampling and upsampling convolutions, and 64 channels for skip convolutions. The upsampling and downsampling modules use $3\times3$ filters, while the skip module uses $1\times1$ filters. In the upsample part of the network, the nearest neighbor algorithm is used. We used $256\times256$ images with three channels in all of our experiments. In training, we use the MSE loss between the degraded image and the output of the network, and we optimize the loss using the ADAM optimizer and a learning rate equal to 0.01 .
In our experiments, we also used $\alpha=0.15$ as an optimal proportion coefficient for reconstruction loss.

\section{Conclusion}
In this work, we introduced a novel framework for contour completion using deep structure priors (DSP). This work offers a novel notion of a maskless grouping of fragmented contours. In our proposed framework, we introduced a novel loss metric that does not require a strict definition of the mask. Instead, it lets the model learn the perceivable illusory contours and connects those fragmented pieces using a generator network that is solely trained on just the single incomplete input image. Our model does not require any pre-training which demonstrates that the convolutional architecture of the hour-glass model is able to connect disconnected contours. We present an extended set of experiments that show the capability of our algorithm. We investigate the effect of each parameter introduced in our algorithm separately and show how one could possibly achieve the best result for their problem using this model. In future work, we plan to extend this model and try to see how it performs with real images. In particular, we want to determine whether we can inpaint real-world photographs while retaining perceptually aware scene structures. The importance of shape in perception by deep neural networks has been highlighted in many adversarial examples to appearance-based networks \cite{baker2020local}. The outcome of this work has strong potential to impact the designing and implementation of models that are robust to such perturbations.

\bibliographystyle{unsrt}  
\bibliography{main}

\end{document}